\documentclass[authoryear,preprint,12pt]{elsarticle}

\usepackage{amssymb}
\usepackage{lineno}
\usepackage{algorithm}
\usepackage{algpseudocode}

\usepackage{caption}
\usepackage{subcaption}
\usepackage{placeins}
\usepackage{setspace}
\usepackage{bbm}
\usepackage{amsfonts}
\usepackage{url}

\journal{Pattern Recognition}

\begin{document}

\begin{frontmatter}

\title{TransductGAN: a Transductive Adversarial Model for Novelty Detection}
\author[label1]{Najiba Toron\corref{cor1}}
\ead{n.toron@cs.ucl.ac.uk}

\author[label1]{Janaina Mourao-Miranda}
\ead{J.Mourao-Miranda@cs.ucl.ac.uk}
\author[label1]{John Shawe-Taylor}
\ead{jst@cs.ucl.ac.uk}

\cortext[cor1]{Corresponding author}
\address[label1]{Department of Computer Science, University College London, London WC1V 6LJ, United Kingdom}

\begin{abstract}
Novelty detection, a widely studied problem in machine learning, is the problem of detecting a novel class of data that has not been previously observed. A common setting for novelty detection is inductive whereby only examples of the negative class are available during training time. Transductive novelty detection on the other hand has only witnessed a recent surge in interest, it not only makes use of the negative class during training but also incorporates the (unlabeled) test set to detect novel examples. Several studies have emerged under the transductive setting umbrella that have demonstrated its advantage over its inductive counterpart. Depending on the assumptions about the data, these methods go by different names (e.g. transductive novelty detection, semi-supervised novelty detection, positive-unlabeled learning, out-of-distribution detection). With the use of generative adversarial networks (GAN), a segment of those studies have adopted a transductive setup in order to learn how to generate examples of the novel class. In this study, we propose TransductGAN, a transductive generative adversarial network that attempts to learn how to generate image examples from both the novel and negative classes by using a mixture of two Gaussians in the latent space. It achieves that by incorporating an adversarial autoencoder with a GAN network, the ability to generate examples of novel data points offers not only a visual representation of novelties, but also overcomes the hurdle faced by many inductive methods of how to tune the model hyperparameters at the decision rule level. Our model has shown superior performance over state-of-the-art inductive and transductive methods.  Our study is fully reproducible with the code available publicly.

\end{abstract}



%%\item we propose a novel architecture for novelty detection capable of %%learning how to generate inliers and novelties simultaneously using distinct %%distributions in the latent space
%%\item by being able to generate the novel points artificially we can train a %%binary classifier and tune the hyperparameters without explicitly  accessing %%the actual outliers; this overcomes the hurdle faced by many inductive %%novelty detection methods of how to tune their hyperparameters in the %%absence of novel data examples
%%\item the introduction of the adversarial autoencoder allows us to use more %%informative features from the latent space when training a binary classifier
%%\item our model exhibits superior performance over several state-of-the-art %5inductive models, and beats a vanilla model that is closely related to a %%state-of-the-art transductive method
%%\item full reproducibility of our model is available on github



\end{frontmatter}

%%\linenumbers

\section{Introduction}

Novelty detection has been addressed widely in the machine learning literature, its basic form aims at building a decision rule in order to distinguish a normal set of data points (or inliers/ negative examples) from novel patterns (or unseen/ positive examples/ novelties). At training time only examples of inliers are available, this type of learning goes by the name of inductive novelty detection. Labeled examples of novelties are unavailable during training in this case. Some notable contributions in this field include one-class classifiers that attempt to draw a discriminative boundary around the normal data such as one-class support vector machine (OCSVM) \citep{scholkopf} and deep learning related methods - especially with the advancements of adversarial learning \citep{goodfellow} - that rely on the fidelity of recontruction of the original signal \citep{schlegl,zenati,zenati2,akcay,akcay2}. One limitation of those inductive methods is their need for hyperparameter tuning; the absence of novel examples in this setup during training makes this problem hard to overcome. 

The traditional inductive learning seeks to infer a decision function by using a training set and minimizing the error rate on the whole distribution of examples. Transductive learning \citep{vapnik} on the other hand is motivated by the idea that the latter approach is unnecessarily complex and that we only care about the decision function so far as the test set is concerned, and hence the (unlabeled) test set should be incorporated at training time. Novelty detection methods that adopt the latter approach are referred to as transductive, one early example adopting this setting is semi-supervised novelty detection introduced by \citep{blanchard10a}.

In this study we propose TransductGAN, a novel transductive novelty detection method using Generative Adversarial Network (GAN) \citep{goodfellow}. Our model combines an adversarial autoencoder \citep{DBLP:journals/corr/MakhzaniSJG15} that maps the data to a latent space with a GAN. It attempts to simultaneously separate the inliers from the novelties in the latent space and learns how to generate both categories of data using two distinct distributions. A standard binary classifier can then be trained using the artificially generated novel samples along with the inlier set in order to perform anomaly detection. A diagram of TransductGAN is provided in Figure \ref{TransductGAN}.

 We refer to the proportion of novelties present in the unlabeled set as the contamination rate. TransductGAN and more generally transductive novelty detection methods are unsuitable for online applications as they assume access to a complete test set. Applications of interest include areas where we benefit from looking at data in a post-event format, and where the novelties are not rare as in the case of anomaly detection (although we will demonstrate empirically that our model outperforms the inductive state-of-the-art methods with a contamination rate as small as 5\%). An example of applications could be the emergence of new topics or themes on social media platforms, such discoveries could help drive the news coverage on more standard media channels. Another application could be in commercial deep neural network systems that are used for classification, and where pre-processing of data is required to maintain performance quality of pre-trained models. This can be achieved by detecting novelties in test data that were not present when models were initially trained (i.e. by monitoring the emergence of a new class). Another example is the discovery of new user profiles that have become interested in acquiring a certain product, this could help boost targeted advertising. In this study we focus on image data.

The main contributions of our work are:
\begin{itemize}
    \item a novel architecture capable of learning how to generate inliers and novelties simultaneously using distinct distributions in the latent space
    \item by being able to generate the novel points artificially we can train a binary classifier and tune the hyperparameters without explicitly  accessing the actual outliers, this overcomes the hurdle faced by many inductive methods of how to tune their hyperparameters in the absence of novel data examples
    \item the introduction of the adversarial autoencoder allows us to use more informative features from the latent space when training a binary classifier
    \item our model exhibits superior performance over several state-of-the-art inductive models, and outperforms a vanilla model that is closely related to a state-of-the-art transductive method
    \item full reproducibility of our model is available on github
\end{itemize}

\section{Background}

\subsection{Wasserstein generative adversarial networks with gradient penalty}

The generative adversarial network (GAN) in its original form \citep{goodfellow} comprises a generator taking as input a latent vector to produce an image output. This network competes against a discriminator tasked with identifying the fake images (produced by the generator) from the real images. By being able to produce samples identical to the real ones at convergence, the generator implicitly captures the distribution of the real image data. Although a GAN can produce visually appealing images, they are hard to train and can suffer from convergence failure or can generate poor quality images. 

Subsequent work \citep{DBLP:journals/corr/SalimansGZCRC16,DBLP:journals/corr/MetzPPS16,arjovsky2017wasserstein,DBLP:journals/corr/PooleASA16,DBLP:journals/corr/abs-1803-05573} has been devoted to boosting training stability. In their work on Wasserstein GAN \citep{arjovsky2017wasserstein}, the authors highlighted that the divergences that are being minimized in the original GAN framework may not be continuous with respect to the generator's parameters which might explain the training difficulty. They instead use the earth mover distance and argue that under mild assumptions this type of loss is continuous and almost differentiable everywhere. They propose a weight clipping of the discriminator in order to enforce those assumptions. The work was further extended with the introduction of Wasserstein GAN with gradient penalty (WGAN-GP) in \citep{DBLP:journals/corr/GulrajaniAADC17} where a gradient penalty was introduced as an alternative to the weight clipping; in this work the authors propose to penalize the norm of the gradient of the discriminator with respect to its input and demonstrate this strategy yields an enhanced performance.

\subsection{Adversarial Autoencoders}
The adversarial autoencoder (AAE) \citep{DBLP:journals/corr/MakhzaniSJG15} is a regularized autoencoder where an adversarial network is tasked with ensuring that the latent code vector distribution is matched with a prior distribution that is defined by the user. The AAE includes an encoder-decoder network that maps the data from the input space to a latent space and a discriminator network that controls the distribution in the latent space. Two losses are used to train the model, the traditional reconstruction based loss that operates in the input space and an adversarial loss that operates on the latent code. The advantage of the AAE is its ability to generate meaningful samples from any part of the latent distribution and its superiority in that regard compared to more traditional methods such as variational autoencoders \citep{kingma2014autoencoding} has been demonstrated.

\section{Related work}

\subsection{Inductive methods}
During training, inductive methods in novelty detection rely on the availability of a negative dataset \( X_n \) that could be either pure or contaminated with novelties or outliers. In our study we assume that the negative dataset is always uncontaminated. There exists a variety of anomaly detection algorithms and comprehensive literature reviews presenting those algorithms as part of a taxonomy already exist, a few examples include \citep{PIMENTEL2014215,DBLP:journals/corr/abs-1901-03407}. The approaches can broadly be categorised into probabilistic models, distance-based, reconstruction-based, domain-based and information theory-based models. We will compare our methods against state-of-the-art models that are either reconstruction-based or domain-based. The main challenge with the other categories is their poor performance on complex high-dimensional data. 

Some notable contributions with regards to the domain-based model include OCSVM \citep{scholkopf}, it attempts to find the boundary enclosing the inlier class by separating it from the origin using a hyperplane with maximum margin. 

The reconstruction-based models rely on modelling the data samples in a latent, lower-dimensional space and on the ability to reconstruct the original samples thereafter. The novelty measure relies on the fidelity of reconstruction, i.e. it relies on some similarity measure between the original sample and its reconstructed version. Algorithms that are able to perform such tasks include neural networks and subspace-based methods. The main advantages of using such models is their scalability, with a large enough sample size they tend to surpass the more traditional methods. 

One state-of-the-art method is AnoGAN, introduced by Schlegl \textit{et al.} in \citep{schlegl}. It uses a deep convolutional GAN (DCGAN, \citep{radford2015unsupervised}) to perform anomaly detection using imaging data. It trains the DCGAN using the inlier data only in the first phase. In the second phase it learns a mapping from the image space to the latent space for each sample in the test set iteratively via backpropagation, using the weighted sum of a residual loss and a discrimination loss. AnoGAN's main contribution is the anomaly scoring that involves mapping from the image to the latent space; it remains a computationally expensive procedure that also requires parameter tuning. 

Efficient GAN-based anomaly detection (EGBAD) \citep{zenati} was introduced soon after and proposes to use a bi-directional GAN (BiGAN) \citep{DBLP:journals/corr/DonahueKD16,dumoulin2016adversarially} in order to overcome the costly process in AnoGAN of recovering the latent representation for a given input. BiGAN simultaneously learns an encoder that maps an input sample to a latent representation along with a generator and discriminator using inlier data. The discriminator takes as input the image sample (either generated or from training) along with its latent representation (either the encoder output or the generator input). With the exception of the introduction of the BiGAN model that allows to reduce the computational costs, the procedure to derive the anomaly score is similar to that of AnoGAN with improved performance. The hyperparameter tuning is still an open issue. 

The GANomaly model was introduced in \citep{akcay} and is inspired by the AAE network. GANomaly employs a variation of this model that uses an added encoder and where the discriminator now operates over the image space. The objectives of GANomaly are similar to the ones presented in the previous two approaches, namely it attempts to use the inlier set to train the model and uses an anomaly score to identify the anomalous samples within the (contaminated) test data. Three types of losses are used to train the model. Firstly, an adversarial loss measures the \(L_2\) distance between the feature representation in the discriminator (hence using the feature matching loss) of the original (image) inputs and the generated images. Secondly, a contextual loss measures the \(L_1\) distance between the input and the generated output and allows the generator to learn contextual information about the input data. And finally an encoder loss minimizes the distance between the latent representation of the input and the latent representation of the generated output. The overall objective function becomes a weighted sum of the three losses. During testing only the encoder loss is used by the model as an novelty indicator. GANomaly's performance was tested on a variety of real-world data and has surpassed other state-of-the-art methods such as EGBAD, AnoGAN as well as more basic methods such as variational autoencoders-based anomaly detection \citep{An2015VariationalAB}.  It should be mentioned that the weights of the objective function were tuned to yield an optimal performance, and in that regard the labels of the test set were used to drive the performance up.

\subsection{Transductive methods}
Transductive methods for novelty detection were originally introduced in \citep{Scott08transductiveanomaly} as a special case of their semi-supervised novelty detection algorithm (SSND) \citep{blanchard10a} where the unlabeled set is incorporated into the the task and where the end goal is to predict the labels of the unlabeled (or test) set. In that study the problem is reduced to a binary classification task subject to a user-specified constraint on the false positive rate. More recently, Tifrea \textit{et al.} introduced ensembles with regularized disagreement (ERD) \citep{DBLP:journals/corr/abs-2012-05825} by training multiple multi-class classifiers on the union of labeled and unlabeled sets. The unlabeled set is assigned random labels and the disagreement between the classifiers is used as a measure of novelty. The method assumes no knowledge of the contamination rate, however the class labels of the labeled set are assumed to be available (i.e. information about the class labels beyond the fact they are inliers). Other transductive state-of-the-art novelty detection methods that make use of the inlier class labels include \citep{DBLP:journals/corr/abs-1908-04951} where the authors train a two-head deep convolutional neural network (CNN) with one common feature extractor and two classifiers; the discrepancy of prediction between the two classifiers is used at test time to separate the inliers from the novel data. The authors have not provided any implementation for this method.

\subsection{PU learning methods}

PU learning refers to the classification of unlabeled samples given data from one class during training. A recent survey of PU learning methods can be found in \citep{DBLP:journals/corr/abs-1811-04820}. The class that is usually available in this setup is the novel/ positive class, but PU learning can also be applied to novelty detection in an unlabeled dataset with access to only the negative data during training, and some studies \citep{pmlr-v37-plessis15} consider transductive novelty detection as defined in \citep{blanchard10a} to be a special case of PU learning. 

The unlabeled data is usually handled in to ways, the first set of methods \citep{Liu02partiallysupervised,10.5555/1630659.1630746} attempt to identify negative examples in the unlabeled set before training a binary classifier using the positive and newly identified negative samples. This approach is very much dependent on heuristic strategies and does not guarantee an optimal solution. The second set of methods \citep{10.5555/951949.952139,10.5555/3041838.3041895} treat the unlabeled set as negative with a weight assigned to each sample, finding the optimal weight is usually costly and the classifiers trained on those datasets suffer from an estimation bias. Several methods have since emerged that seek to build an unbiased PU classifier \citep{10.5555/2968826.2968905,pmlr-v37-plessis15,kiryo2017positiveunlabeled}, the most recent of which, the non-negative PU estimator (nnPU) \citep{kiryo2017positiveunlabeled}, was proven to be the most robust against overfitting in the presence of a flexible classifier. More recently, GAN-based models have become state-of-the-art in PU learning applications and a couple of methods are closely related to our model as they also attempt to generate counter-examples. In \citep{DBLP:journals/corr/abs-1711-08054} a generative positive-unlabeled (GenPU) model is proposed that makes use of a series of discriminators and generators to produce both positive and negative samples; GenPU requires prior knowledge of the contamination rate in the unlabeled set. The authors have not provided any implementation of that model. Following this work, Chiaroni \textit{et al.} have proposed a two-stage GAN model (D-GAN) that learns the counter-examples distribution before training a binary classifier using the fake generated negative examples and the already available positive examples. D-GAN incorporates a biased PU risk in the discriminator loss function that constrains the generator to learn the positive samples distribution exclusively. The study also demonstrates that the standard GAN loss function in use also alleviates the need of prior knowledge of the contamination rate.

\section{TransductGAN}

\begin{figure}
  \centering
  \includegraphics[scale=0.6]{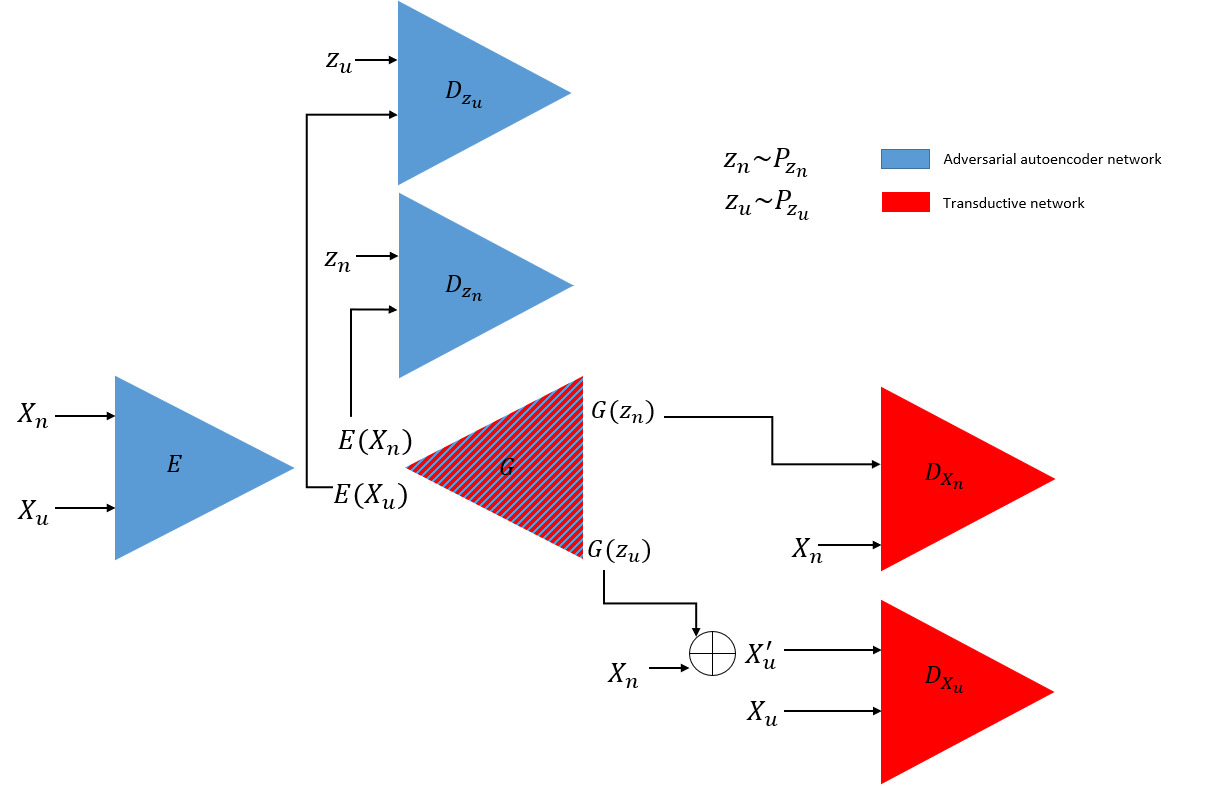}
  \caption{TransductGAN. \textit{The network includes an adversarial autoencoder that ensures the latent projection's distributions match with the priors as per equation \ref{eq:p_u}, and a transductive network that learns to map samples from \(p_n(\textbf{z})\) and \(p_p(\textbf{z})\) into negative and positive examples respectively. The two networks are trained simultaneously. The training of \(D_{X_n}\) is optional and only needed if generating samples from the negative data is required.}}\label{TransductGAN}
\end{figure}

\subsection{Model objectives}

TransductGAN's objectives are to be able to learn two distinct distributions, one (optionally) for the inlier data and another for the novel data while having access to a negative dataset and an unlabeled dataset. It is able to achieve that by making use of a latent space. TransductGAN is a transductive adversarial network that includes an adversarial autoencoder that matches the aggregated posterior distribution of the latent space with a bimodal distribution, and a generative model that learns how to map each mode in the latent space to either positive or negative data outputs. By being able to separate the projections of the positive and negative data in the latent space, TransductGAN is able to generate both novel and negative data separately. A binary classifier (e.g. a support vector machine) can then be trained to identify the novel data points in the unlabeled set. This operation takes place in the absence of any labeled positive data examples.

More formally, let us define the input as \(x \in \mathbbm{R}^m\) and its projection in the latent space (i.e. the output of the encoder \(E\) in Figure \ref{TransductGAN}) as \(z \in \mathbbm{R}^{n}\) with \(m>>n\). We have access to the negative dataset \(X_n=\{x_n^i\}_{i=1}^{n_n} \sim P_{x_n}\) and an unlabeled dataset that includes novel data \(X_u=\{x_u^i\}_{i=1}^{n_u} \sim P_{x_u}\). Let us also define \(y \in \{-1,+1\}\) as the class label (novel or negative). The class conditional positive density (representing the positive samples in the latent space) is:

\begin{equation}
    q_p(z)=q(z|y=1),
\end{equation}
and the class conditional negative density (representing the negative data samples) in the latent space:
\begin{equation}
    q_n(z)=q(z|y=-1).
\end{equation}

\noindent The unlabeled density in the latent space is hence defined as:
\begin{equation}
    q_u(z)=\pi q_p(z) + (1-\pi) q_n(z).
\end{equation}

\noindent \(\pi\) is the prior for the positive class also known as the contamination rate, we assume we have access to this figure and that it is strictly positive \(\pi>0\):

\begin{equation}
    \pi=p(y=1)
\end{equation}

\noindent The prior distribution we want to impose directly are \(p_n(z)\) for the negative dataset with \(N(\mu_n,\Sigma_n)\) and \(p_u(z)\) for the unlabeled dataset, and indirectly \(p_p(z)\) for the positive dataset with \(N(\mu_p,\Sigma_p)\). The unlabeled prior can also be defined as:

\begin{equation} \label{eq:p_u}
    p_u(z)=\pi p_p(z) + (1-\pi)p_n(z)
\end{equation}

\noindent We assume we have access to \(n_n\) negative samples and \(n_u\) unlabeled samples.

\subsection{Training methods}
The proposed model for learning how to generate negative and positive samples is summarised in Figure \ref{TransductGAN}. The first part of the model is an adversarial autoencoder that includes the networks \(E\), \(G\),  and the critic networks \(D_{z_u}\) and \(D_{z_n}\). \(E\) and \(G\) are trained to minimize a reconstruction loss, or a minimal \(L_2\) distance between an image and its reconstructed version:

\begin{equation}
    L_{reconstruction}=\min_{E} \min_{G} \mathbb{E}_{x \sim P_x} \|x-G(E(x))\|_2
\end{equation}
\noindent In the above \(P_{x}\) could refer to either \(P_{X_u}\) or \(P_{X_n}\).

When using \(X_u\), \(D_{z_u}\) and \(E\) are trained  to match the projections of \(X_u\) onto the latent space with \(p_u(z)\) defined in equation \ref{eq:p_u}, a Wasserstein loss with a gradient penalty (WGAN-GP) \citep{DBLP:journals/corr/GulrajaniAADC17} is used in order to ensure that. And similarly when using \(X_n\), \(D_{z_n}\) and \(E\) are trained to match the projections of \(X_n\) onto the latent space with \(p_n(z)\). In both cases, the latent loss can be defined as:

\begin{equation}
    L_{regularization}=\min_{E} \max_{D} \mathbb{E}_{x \sim P_x}[D(E(x))]-\mathbb{E}_{z \sim p(z)}D(z)+\lambda \mathbb{E}_{\hat{z} \sim P_{\hat{z}}}[(\|\nabla_{\hat{z}} D(\hat{z})\|_2 - 1)^2]
\end{equation}

\noindent In the above, depending on whether we are using \(X_u\) or \(X_n\) during training, \(P_{x}\) refers to either \(P_{X_u}\) or \(P_{X_n}\). \(p(z)\) refers to either \(p_u(z)\) or \(p_n(z)\) and the critic network \(D\) refers to either \(D_{z_u}\) or \(D_{z_n}\). Similarly to the gradient penalty definition in \citep{DBLP:journals/corr/GulrajaniAADC17}, the sampling  \(P_{\hat{z}}\) is a uniform distribution along straight lines between pairs of points sampled from \((q_u(z),p_u(z))\) or \((q_n(z),p_n(z))\). By imposing \(p_u(z)\) and \(p_n(z)\) on \(q_u(z)\) and \(q_n(z)\) respectively, \(q_p(z)\) will implicitly match with \(p_p(z)\).

The second part of the model focuses on \(G\), \(D_{X_u}\) and \(D_{X_n}\). The generator \(G\) will be trained to produce fake positive samples \(G(z_p)\), and (optionally) fake negative samples \(G(z_n)\). Let us define \(X_{u}^{'}\) to be the result of a concatenation operation between \(X_{n}\) and \(G(z_p)\) with proportions of \(1- \pi\) and \(\pi\) respectively. With regards to generating \(G(z_p)\), the intuition is that \(X_u\) is formed by samples from both the negative set \(X_n\) and unknown positive samples \(X_p\), we ensure \(X_n\) already forms a subset of \(X_u^{'}\), the generator \(G\) is then left with the task of complementing \(X_u^{'}\) with samples not already provided by \(X_n\), i.e. samples that are similar to the novel samples \(X_p\). The role of \(D_{x_u}\) is to challenge \(G\) to produce a set \(G(z_p)\) that is indistinguishable from \(X_p\). 

With regards to generating \(G(z_n)\), this follows a standard GAN training approach where \(G\) is trained to generate fake negative samples that should look like \(X_n\), and \(D_{X_n}\) is trained to challenge \(G\) on that task.

The adversarial loss, also a WGAN-GP, is defined as:
\begin{equation}
    L_{adv}=\min_{G} \max_{D} \mathbb{E}[D(X^{'})]-\mathbb{E}[D(X)]+\lambda \mathbb{E}_{\hat{x} \sim P_{\mathbf{\hat{x}}}}[(\|\nabla_{\hat{x}} D(\mathbf{\hat{x}})\|_2 - 1)^2]
\end{equation}

\noindent In the above, depending on whether we are using \(z_u\) or \(z_n\) during training, \(X^{'}\) refers to either \(X_u^{'}\) or \(G(z_n)\). \(X\) refers to either \(X_u\) or \(X_n\) and the critic network \(D\) refers to either \(D_{X_u}\) or \(D_{X_n}\). Similarly to the gradient penalty definition in \citep{DBLP:journals/corr/GulrajaniAADC17}, the sampling  \(P_{\hat{x}}\) is a uniform distribution along straight lines between pairs of points sampled from \((X_{u}^{'},X_u)\) or \((G(z_n),X_n)\). A pseudocode of TransductGAN is provided in algorithm \ref{alg:TransductGAN}.

Once the TransductGAN training is completed, a binary classifier can be trained using the latent projections of \(X_n\) onto the latent space as one class and the latent projections of the fake positive images as the other class as per Algorithm \ref{alg:novelty_detector}.

\begin{algorithm} 
\caption{TransductGAN. We use default values of \(\lambda=10\), \(n_{critic}=5\) \citep{DBLP:journals/corr/GulrajaniAADC17}.} 
\begin{algorithmic}\label{alg:TransductGAN}
\Require the contamination rate \(\pi\), the gradient penalty coefficient \(\lambda\), the number of critic iterations per generator iteration \(n_{critic}\), the batch size \(m\).
\Require initial critics parameters \(\theta_{D_{z_u}}\), \(\theta_{D_{z_n}}\), \(\theta_{D_{X_u}}\), \(\theta_{D_{X_n}}\), initial encoder parameters \(\theta_{E}\) and initial generator parameters \(\theta_{G}\).
\State sample \(m\) samples from \(X_u\)
\State minimize \(L_{rec}\) wrt \(\theta_{E}\) and \(\theta_{G}\) and update parameters accordingly
\State sample \(z \sim p_{u}(z)\)
\State minimize \(L_{reg}\) wrt \(\theta_{E}\) and update parameters accordingly
\For {\(i=1,...,n_{critic}\)}
\State maximize \(L_{reg}\) wrt \(\theta_{D_{z_u}}\) and update parameters accordingly
\EndFor
\State sample \(m\) samples from \(X_n\)
\State minimize \(L_{rec}\) wrt \(\theta_{E}\) and \(\theta_{G}\) and update parameters accordingly
\State sample \(z \sim p_{n}(z)\)
\State minimize \(L_{reg}\) wrt \(\theta_{E}\) and update parameters accordingly
\For {\(i=1,...,n_{critic}\)}
\State maximize \(L_{reg}\) wrt \(\theta_{D_{z_n}}\) and update parameters accordingly 
\EndFor
\State sample \(m\) samples from \(X_u\)
\State sample \(int(\pi*m\)) samples \(z \sim p_{p}(z)\)
\State combine \(G(z)\) with \((m-int(\pi*m))\) samples from \(X_{n}\) to form \(X_{u}^{'}\)
\State minimize \(L_{adv}\) wrt \(\theta_{G}\) and update parameters accordingly
\For {\(i=1,...,n_{critic}\)}
\State maximize \(L_{adv}\) wrt \(\theta_{D_{X_{u}}}\) and update parameters accordingly
\EndFor
\State sample \(m\) samples from \(X_n\)
\State sample \(z \sim p_{n}(z)\)
\State minimize \(L_{adv}\) wrt \(\theta_{G}\) and update parameters accordingly
\For {\(i=1,...,n_{critic}\)}
\State maximize \(L_{adv}\) wrt \(\theta_{D_{X_n}}\) and update parameters accordingly
\EndFor

\end{algorithmic}
\end{algorithm}

\begin{algorithm}
\caption{Binary classifier for novelty detection.
}\label{alg:novelty_detector}
\begin{algorithmic}
\Require Training of TransductGAN as per Algorithm \ref{alg:TransductGAN}
\State sample \(5000\) samples \(z_p\) from \( p_{p}(z)\)
\State sample \(5000\) samples \(X_n\) from \(X_n\)
\State train a two-class SVM with linear kernel with \(E(G(z_p))\) as one class and \(E(X_n)\) as another
\State apply classifier on \(E(X_u)\) as novelty detector
\end{algorithmic}
\end{algorithm}

\begin{algorithm} 
\caption{Vanilla model. We use default values of \(\lambda=10\), \(n_{critic}=5\) \citep{DBLP:journals/corr/GulrajaniAADC17}.}\label{alg:vanilla}
\begin{algorithmic}
\Require the contamination rate \(\pi\), the gradient penalty coefficient \(\lambda\), the number of critic iterations per generator iteration \(n_{critic}\), the batch size \(m\).
\Require initial critics parameters \(\theta_{D_{X_u}}\) and initial generator parameters \(\theta_{G}\).

\State sample \(m\) samples from \(X_u\)
\State sample \(int(\pi*m\)) samples \(z \sim N(\mu, \Sigma)\)
\State combine \(G(z)\) with \((m-int(\pi*m))\) samples from \(X_{n}\) to form \(X_{u}^{'}\)
\State minimize \(L_{adv}\) wrt \(\theta_{G}\) and update parameters accordingly
\For {\(i=1,...,n_{critic}\)}
\State maximize \(L_{adv}\) wrt \(\theta_{D_{X_{u}}}\) and update parameters accordingly
\EndFor
\end{algorithmic}
\end{algorithm}

\begin{algorithm}
\caption{Binary classifier for novelty detection with vanilla model.
}\label{alg:novelty_detector_vanilla}
\begin{algorithmic}
\Require Training of vanilla model as per Algorithm \ref{alg:vanilla}
\State sample \(5000\) samples \(z\) from \(N(\mu,\Sigma)\)
\State sample \(5000\) samples from \(X_n\)
\State train a two-class SVM with radial basis function kernel with \(G(z)\) as one class and \(X_n\) as another
\State apply classifier on \(E(X_u)\) as novelty detector
\end{algorithmic}
\end{algorithm}

\section{Experimental results}
\subsection{Datasets}
\subsubsection{MNIST \citep{lecun}.} This dataset of handwritten digits has a training set of 60000 examples, and a test set of 10000 examples and includes 10 classes. In our novelty detection formulation we will treat one of the classes as novel and remove it from the training set, the rest of the classes will be treated as normal. We will iterate through all the class combinations when reporting our results.

\subsubsection{CIFAR10 \citep{krizhevsky}.} This dataset consists of 50000 training images and 10000 test images of 10 classes. We will proceed as with MNIST regarding our novelty detection formulation. 

\subsection{Performance measure}
Our performance will be measured using the area under the curve of the Receiver Operating Characteristic (AUROC). The Receiver Operating Characteristic curve plots the true positive rate against the false positive rate as we vary the threshold of our classifier. We will also include image outputs of the produced novel images.
\subsection{Network architectures}
The transductive network architecture as highlighted in red in Figure \ref{TransductGAN} follows the same implementation  \footnote{https://github.com/igul222/improved\_wgan\_training} as was provided by \citep{gulrajani2017improved}. Our open source implementation provides further details about the adversarial network implementation. 

\begin{figure}[hbt!]
     \centering
     \begin{subfigure}[b]{0.4\textwidth}
         \centering
         \includegraphics[width=\textwidth]{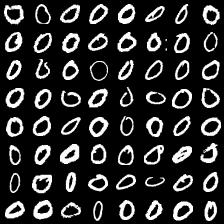}
         \caption{Real novel images}
         \label{fig:mnist0}
     \end{subfigure}
     \hfill
     \begin{subfigure}[b]{0.4\textwidth}
         \centering
         \includegraphics[width=\textwidth]{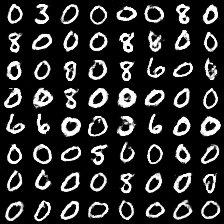}
         \caption{Fake novel images (5\%)}
         \label{fig:mnist1}
     \end{subfigure}
     \hfill
     \begin{subfigure}[b]{0.4\textwidth}
         \centering
         \includegraphics[width=\textwidth]{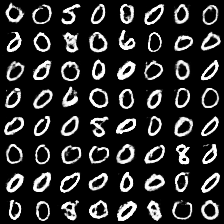}
         \caption{Fake novel images (10\%)}
         \label{fig:mnist2}
     \end{subfigure}
     \hfill
     \begin{subfigure}[b]{0.4\textwidth}
         \centering
         \includegraphics[width=\textwidth]{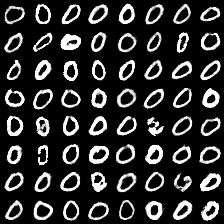}
         \caption{Fake novel images (30\%)}
         \label{fig:mnist3}
     \end{subfigure}
        \caption{MNIST example with '0' as novel class. For the fake examples, the value in brackets corresponds to the contamination rate in the test set that was used during training.}
        \label{fig:mnist4}
\end{figure}

\begin{figure}[hbt!]
     \centering
     \begin{subfigure}[b]{0.4\textwidth}
         \centering
         \includegraphics[width=\textwidth]{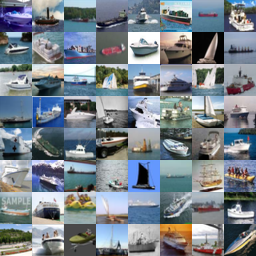}
         \caption{Real novel images}
         \label{fig:cif0}
     \end{subfigure}
     \hfill
     \begin{subfigure}[b]{0.4\textwidth}
         \centering
         \includegraphics[width=\textwidth]{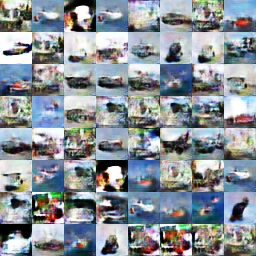}
         \caption{Fake novel images (5\%)}
         \label{fig:cif1}
     \end{subfigure}
     \hfill
     \begin{subfigure}[b]{0.4\textwidth}
         \centering
         \includegraphics[width=\textwidth]{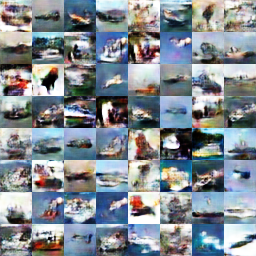}
         \caption{Fake novel images (10\%)}
         \label{fig:cif2}
     \end{subfigure}
     \hfill
     \begin{subfigure}[b]{0.4\textwidth}
         \centering
         \includegraphics[width=\textwidth]{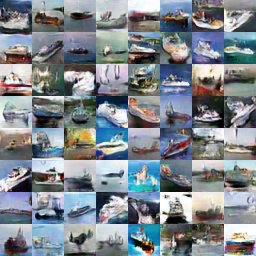}
         \caption{Fake novel images (30\%)}
         \label{fig:cif3}
     \end{subfigure}
        \caption{CIFAR10 example with ship as novel class. For the fake examples, the value in brackets corresponds to the contamination rate in the test set that was used during training.}
        \label{fig:cif4}
    
\end{figure}

\subsection{Methods for comparison}
\begin{figure}
  \centering
  \includegraphics[scale=0.55]{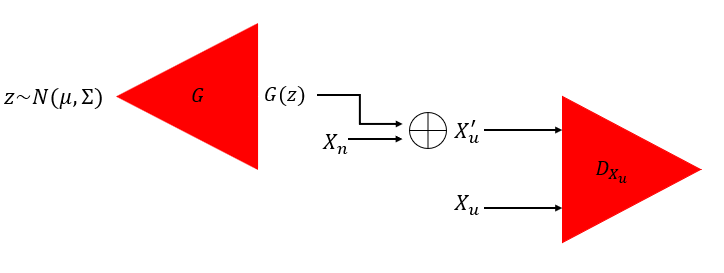}
  \caption{Vanilla transductive model. \textit{The network is only based on a GAN model and does not exploit learning in the latent space as with TransductGAN. The generator will learn to generate positive fake images that will be used along with the negative images \(X_n\) to train a binary classifier. The algorithm is outlined in Algorithm \ref{alg:vanilla} and the classification task is summarised in Algorithm \ref{alg:novelty_detector_vanilla}.}}\label{vanilla}
\end{figure}

We will compare our model against OCSVM, EGBAD and GANomaly. We use the scikit implementation of OCSVM \footnote{https://scikit-learn.org/stable/modules/generated/sklearn.svm.OneClassSVM.html} with the 'scale' default kernel width value. We follow the same training procedures stated in the original publications regarding EGBAD and GANomaly's results. EGBAD's results with CIFAR10 were taken from \citep{akcay}. The D-GAN implementation made available by the authors did not include the version they used with MNIST and CIFAR10 datasets; our attempt to reproduce their results resulted in a mode collapse with classification results no better than random so we have not included these in our comparison. We have however built a vanilla model of TransductGAN that is very close to D-GAN with the exception that it requires prior knowledge of the contamination rate (its performance could hence be seen as a best case scenario of D-GAN), it adopts an architecture that is similar to the transductive TransductGAN in order to ensure a fair comparison. The vanilla network does not make use of a latent space and trains a binary classifier (a SVM with a radial basis function kernel) based on the fake (positive) generated samples from the generator and the negative images \(X_n\), it is summarised in Figure \ref{vanilla}) and an outline of the overall procedure is provided in Algorithm \ref{alg:vanilla} and Algorithm \ref{alg:novelty_detector_vanilla}.

\begin{table}[hbt!]
  \resizebox{1.25\textwidth}{!}{
  %%\begin{minipage}{\textwidth}

  \begin{tabular}{llllllllll}
  
  %%  \toprule
    &OCSVM&GANomaly&EGBAD&TransductGAN(5\%)&Vanilla(5\%)&TransductGAN(10\%)&Vanilla(10\%)&TransductGAN(30\%)&Vanilla(30\%)\\
    \hline
    %%\midrule
0&0.853(0)&0.882&0.86&0.983(0.002)&0.985(0.003)&0.991(0.002)&0.995(0)&0.996(0.001)&0.998(0)\\
1&0.315(0.001)&0.663&0.314&0.994(0.001)&0.997(0)&0.996(0)&0.996(0)&0.998(0)&0.998(0)\\
2&0.775(0.002)&0.952&0.835&0.954(0.001)&0.944(0.009)&0.985(0.002)&0.981(0.001)&0.997(0.002)&0.997(0)\\
3&0.655(0.004)&0.794&0.712&0.949(0.007)&0.955(0.007)&0.975(0.005)&0.972(0.004)&0.996(0.002)&0.995(0.001)\\
4&0.498(0.003)&0.803&0.655&0.952(0.008)&0.948(0.005)&0.978(0.004)&0.976(0.005)&0.994(0.001)&0.997(0)\\
5&0.589(0.003)&0.864&0.713&0.917(0.007)&0.924(0.007)&0.975(0.002)&0.962(0.004)&0.994(0.001)&0.996(0)\\
6&0.691(0.003)&0.852&0.753&0.983(0.003)&0.987(0)&0.994(0.002)&0.994(0)&0.996(0.001)&0.998(0)\\
7&0.582(0.009)&0.697&0.525&0.96(0.005)&0.954(0.003)&0.976(0.005)&0.974(0.001)&0.993(0.001)&0.996(0)\\
8&0.544(0.002)&0.792&0.728&0.936(0.006)&0.921(0)&0.967(0.001)&0.964(0.003)&0.992(0)&0.992(0.001)\\
9&0.349(0.005)&0.534&0.547&0.938(0.012)&0.902(0.018)&0.968(0.005)&0.912(0.063)&0.988(0.002)&0.99(0.001)\\
\hline
    %%\bottomrule
    &0.585&0.783&0.664&0.956&0.952&0.98&0.973&0.994&0.996

  \end{tabular}
  %%\end{minipage}
  }
    \caption{ROC (AUC) results summary - MNIST. Results for TransductGAN are from the z features. Three different random seeds are used with standard deviations shown in brackets for TransductGAN, the vanilla model and OCSVM.}
  \label{tbl-2}
\end{table}

\begin{table} [hbt!]
  \resizebox{1.25\textwidth}{!}{
  %%\begin{minipage}{\textwidth}

  \begin{tabular}{llllllllll}
  
    %%\toprule 
    &OCSVM&GANomaly&EGBAD&TransductGAN(5\%)&Vanilla(5\%)&TransductGAN(10\%)&Vanilla(10\%)&TransductGAN(30\%)&Vanilla(30\%)\\
    \hline
   %% \midrule
plane&0.54(0.01)&0.622&0.582&0.763(0.003)&0.625(0.005)&0.81(0.004)&0.786(0.009)&0.86(0.002)&0.84(0)\\
car&0.657(0.002)&0.632&0.527&0.7(0.009)&0.756(0.018)&0.85(0.002)&0.77(0.007)&0.906(0.004)&0.843(0.037)\\
bird&0.384(0.011)&0.513&0.386&0.6(0.018)&0.39(0.011)&0.66(0.007)&0.633(0.015)&0.766(0.008)&0.713(0.018)\\
cat&0.577(0.012)&0.575&0.455&0.586(0.045)&0.56(0.013)&0.655(0.004)&0.64(0)&0.786(0.009)&0.69(0.025)\\
deer&0.32(0.005)&0.591&0.385&0.55(0.007)&0.376(0.055)&0.686(0.007)&0.62(0.008)&0.796(0)&0.713(0.006)\\
dog&0.57(0.005)&0.625&0.490&0.65(0.021)&0.62(0.01)&0.715(0.011)&0.68(0.018)&0.816(0.006)&0.753(0.004)\\
frog&0.397(0.001)&0.668&0.359&0.613(0.022)&0.493(0.073)&0.736(0.015)&0.643(0.007)&0.86(0.011)&0.746(0.011)\\
horse&0.54(0.005)&0.650&0.527&0.636(0.021)&0.606(0.007)&0.74(0.011)&0.696(0.012)&0.856(0.009)&0.78(0.031)\\
ship&0.51(0.009)&0.622&0.411&0.8(0.007)&0.736(0.008)&0.87(0.006)&0.836(0.019)&0.9(0.001)&0.89(0.003)\\
truck&0.66(0.006)&0.615&0.554&0.696(0.027)&0.77(0.007)&0.81(0.009)&0.776(0.005)&0.876(0.005)&0.816(0.028)\\
\hline
    %%\bottomrule
    &0.515&0.611&0.468&0.6596&0.5935&0.7533&0.7053&0.8426&0.7786

  \end{tabular}
  %%\end{minipage}
  }
    \caption{ROC (AUC) results summary - CIFAR10. Three different random seeds are used with standard deviations shown in brackets for TransductGAN, the vanilla model and OCSVM.}
  \label{tbl-3}
\end{table}

\subsection{Discussion}
The results of our experiments are presented in Table \ref{tbl-2} (MNIST) and Table \ref{tbl-3} (CIFAR10). Each row corresponds to a different novelty class and the last row summarises the average performance in terms of AUROC score.  On average, our TransductGAN model outperforms GANomaly (the best performing inductive model) by 22.09\% (MNIST) and 7.95\% (CIFAR10) when the contamination rate in the test set is 5\%. With a contamination rate of 10\% TransductGAN outperforms GANomaly by 25.15\% (MNIST) and by 23.28\% (CIFAR10). When the latter figure increases to 30\%, TransductGAN outperforms GANomaly by 26.94\% (MNIST) and 37.9\% (CIFAR10). This highlights the suitability of our model in cases where the novelties are not rare, however even in cases where the contamination rate drops to 5\% our model still outperforms GANomaly. Unsurprisingly, the higher the contamination rate the better the performance as the model is able to learn from a wider pool of novel data points. This is further confirmed by assessing the improvement in image quality of the fake novel samples in Figure \ref{fig:mnist3} and Figure \ref{fig:cif3} as the contamination rate increases. It should also be noted that EGBAD and GANomaly's performances rely on hyperparameter tuning as noted in their respective publications, the tuning was carried to yield optimal performance which suggests the test set labels were used in the process. Our model, although transductive, does not make use of the test set labels at any training stage. Finally, our model also offers the advantage of generating examples of good quality novel images as shown in Figure \ref{fig:mnist3} and Figure \ref{fig:cif3}. Some further examples of novel image generation are included in Appendix \ref{appa}. 

As mentioned previously, we were not able to carry any direct comparison to state-of-the-art transductive methods due to the difference in experimental design where we do not make use of the class labels in the training set (as opposed to the work of \citep{DBLP:journals/corr/abs-2012-05825} for example). The state-of-the-art PU learning method (D-GAN) did not have a full implementation available, and although it has an advantage over our method (i.e. the fact it does not require any prior knowledge of the contamination rate in the test set), it only works with a specific type of discriminator loss (the cross-entropy). Previous studies have shown the limitations of using this type of loss and have highlighted issues such as mode collapse or non-informative gradients. Optimal transport type of losses \citep{arjovsky2017wasserstein,DBLP:journals/corr/abs-1803-05573} have shown more stability in the training process, our model has the flexibility of using any of those losses during training. We have compared our method to a vanilla version of TransductGAN that is closely related to D-GAN. Its performance results are displayed under the 'Vanilla' columns (with varying rates of contamination) in Table \ref{tbl-2} and Table \ref{tbl-3}. With regards to MNIST, our model outperforms the vanilla version by 0.51\% and 0.8\% for a contamination rate of 5\% and 10\% respectively, it underperforms the vanilla version by -0.15\% for a contamination rate of 30\%. On average across all contamination rates, it outperforms the vanilla model by 0.39\%. This low performance comparison level is due to the MNIST results already being \(>0.95\) for the vast majority. With regards to CIFAR10, our model outperforms the vanilla version by 10.03\%, 6.37\% and 7.59\% for a contamination rate of 5\%, 10\% and 30\% respectively, this demonstrates a superior performance across all contamination levels. 

\section{Conclusion and future work}
We have introduced a novel architecture that learns to generate inliers and novel data using separate distributions by combining an adversarial autoencoder with a GAN under a transductive mode. A binary classifier is trained in the second stage using the latent features of the encoder network to distinguish the novel samples from the inliers. By being able to generate the novel data points artificially we overcome the hurdle of hyperparameter tuning that is faced by state-of-the-art inductive methods. Our model outperforms the latter methods and is capable of reproducing good quality visual representations of the novel images. Our model also outperforms a vanilla model that is closely related to D-GAN; this highlights the advantage that the latent space projections offer in TransductGAN over the image space features in the vanilla model. TransductGAN is most suitable for situations where the novelties are not rare but also shows competitive results when the contamination rate drops to low levels. It also requires prior knowledge of the contamination rate which is a disadvantage over other state-of-the-art PU learning methods, but it allows for the use of more powerful loss functions with enhanced stability during training. Future work will focus on learning the contamination rate and assessing cases where the contamination rate falls to levels below 5\%. Future work will also extend to include labels of the inlier set during training when these are available; this should yield an enhanced performance and allow us to compare our model against a wider range of transductive models. 

\section*{Acknowledgements}

\noindent JMM is supported by the Wellcome Trust under Grant No. WT102845/Z/13/Z. NT is funded by the Engineering and Physical Sciences Research Council. We would like to thank Giulia Luise and Florent Chiaroni for insightful comments.




\appendix
\section{}
\label{appa}
This appendix presents additional results using different novel class examples with MNIST and CIFAR10 datasets.

\begin{figure}[!htbp]
     \centering
     \begin{subfigure}[b]{0.4\textwidth}
         \centering
         \includegraphics[width=\textwidth]{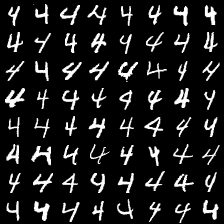}
         \caption{Real novel images}
         \label{fig:mnist0_4}
     \end{subfigure}
     \hfill
     \begin{subfigure}[b]{0.4\textwidth}
         \centering
         \includegraphics[width=\textwidth]{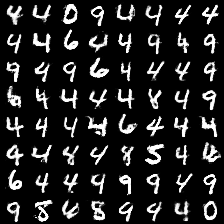}
         \caption{Fake novel images (5\%)}
         \label{fig:mnist1_4}
     \end{subfigure}
     \hfill
     \begin{subfigure}[b]{0.4\textwidth}
         \centering
         \includegraphics[width=\textwidth]{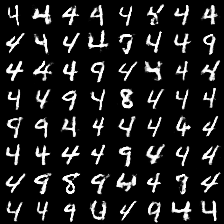}
         \caption{Fake novel images (10\%)}
         \label{fig:mnist2_4}
     \end{subfigure}
     \hfill
     \begin{subfigure}[b]{0.4\textwidth}
         \centering
         \includegraphics[width=\textwidth]{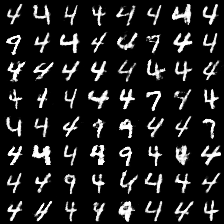}
         \caption{Fake novel images (30\%)}
         \label{fig:mnist3_4}
     \end{subfigure}
        \caption{MNIST example with '4' as novel class. For the fake examples, the value in brackets corresponds to the contamination rate in the test set that was used during training.}
        \label{fig:mnist4_4}
\end{figure}
\FloatBarrier

\begin{figure}[!htbp]
     \centering
     \begin{subfigure}[b]{0.4\textwidth}
         \centering
         \includegraphics[width=\textwidth]{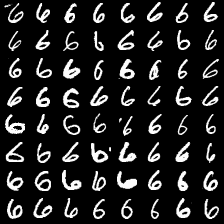}
         \caption{Real novel images}
         \label{fig:mnist0_6}
     \end{subfigure}
     \hfill
     \begin{subfigure}[b]{0.4\textwidth}
         \centering
         \includegraphics[width=\textwidth]{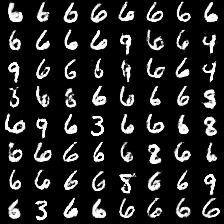}
         \caption{Fake novel images (5\%)}
         \label{fig:mnist1_6}
     \end{subfigure}
     \hfill
     \begin{subfigure}[b]{0.4\textwidth}
         \centering
         \includegraphics[width=\textwidth]{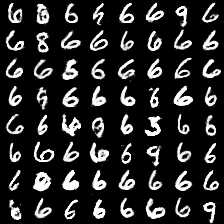}
         \caption{Fake novel images (10\%)}
         \label{fig:mnist2_6}
     \end{subfigure}
     \hfill
     \begin{subfigure}[b]{0.4\textwidth}
         \centering
         \includegraphics[width=\textwidth]{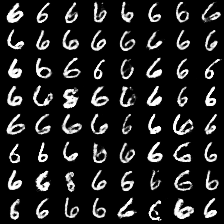}
         \caption{Fake novel images (30\%)}
         \label{fig:mnist3_6}
     \end{subfigure}
        \caption{MNIST example with '6' as novel class. For the fake examples, the value in brackets corresponds to the contamination rate in the test set that was used during training.}
        \label{fig:mnist4_6}
\end{figure}
\FloatBarrier

\begin{figure}[!htbp]
     \centering
     \begin{subfigure}[b]{0.4\textwidth}
         \centering
         \includegraphics[width=\textwidth]{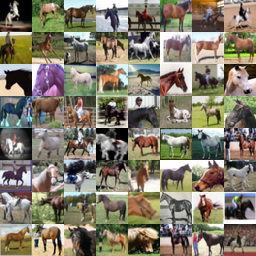}
         \caption{Real novel images}
         \label{fig:cif0_horse}
     \end{subfigure}
     \hfill
     \begin{subfigure}[b]{0.4\textwidth}
         \centering
         \includegraphics[width=\textwidth]{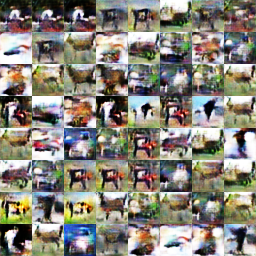}
         \caption{Fake novel images (5\%)}
         \label{fig:cif1_horse}
     \end{subfigure}
     \hfill
     \begin{subfigure}[b]{0.4\textwidth}
         \centering
         \includegraphics[width=\textwidth]{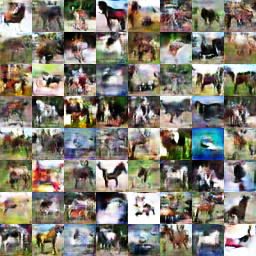}
         \caption{Fake novel images (10\%)}
         \label{fig:cif2_horse}
     \end{subfigure}
     \hfill
     \begin{subfigure}[b]{0.4\textwidth}
         \centering
         \includegraphics[width=\textwidth]{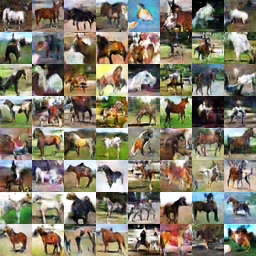}
         \caption{Fake novel images (30\%)}
         \label{fig:cif3_horse}
     \end{subfigure}
        \caption{CIFAR10 example with horse as novel class. For the fake examples, the value in brackets corresponds to the contamination rate in the test set that was used during training.}
        \label{fig:cif4_horse}
\end{figure}
\FloatBarrier

\begin{figure}[!htbp]
     \centering
     \begin{subfigure}[b]{0.4\textwidth}
         \centering
         \includegraphics[width=\textwidth]{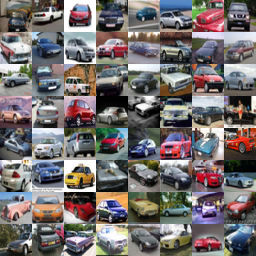}
         \caption{Real novel images}
         \label{fig:cif0_car}
     \end{subfigure}
     \hfill
    \begin{subfigure}[b]{0.4\textwidth}
         \centering
         \includegraphics[width=\textwidth]{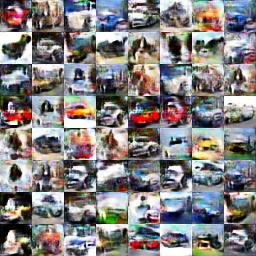}
         \caption{Fake novel images (5\%)}
         \label{fig:cif1_car}
     \end{subfigure}
     \hfill
     \begin{subfigure}[b]{0.4\textwidth}
         \centering
         \includegraphics[width=\textwidth]{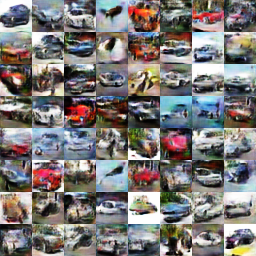}
         \caption{Fake novel images (10\%)}
         \label{fig:cif2_car}
     \end{subfigure}
     \hfill
     \begin{subfigure}[b]{0.4\textwidth}
         \centering
         \includegraphics[width=\textwidth]{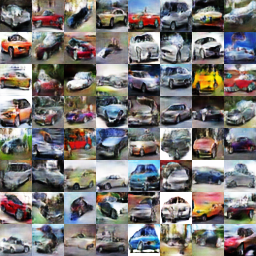}
         \caption{Fake novel images (30\%)}
         \label{fig:cif3_car}
     \end{subfigure}
        \caption{CIFAR10 example with car as novel class. For the fake examples, the value in brackets corresponds to the contamination rate in the test set that was used during training.}
        \label{fig:cif4_car}
\end{figure}

\bibliographystyle{elsarticle-harv.bst}
\bibliography{bib.bib}

\begin{thebibliography}{38}
\expandafter\ifx\csname natexlab\endcsname\relax\def\natexlab#1{#1}\fi
\providecommand{\url}[1]{\texttt{#1}}
\providecommand{\href}[2]{#2}
\providecommand{\path}[1]{#1}
\providecommand{\DOIprefix}{doi:}
\providecommand{\ArXivprefix}{arXiv:}
\providecommand{\URLprefix}{URL: }
\providecommand{\Pubmedprefix}{pmid:}
\providecommand{\doi}[1]{\href{http://dx.doi.org/#1}{\path{#1}}}
\providecommand{\Pubmed}[1]{\href{pmid:#1}{\path{#1}}}
\providecommand{\bibinfo}[2]{#2}
\ifx\xfnm\relax \def\xfnm[#1]{\unskip,\space#1}\fi
%Type = Article
\bibitem[{Akcay et~al.(2018)Akcay, Abarghouei and Breckon}]{akcay}
\bibinfo{author}{Akcay, S.}, \bibinfo{author}{Abarghouei, A.A.},
  \bibinfo{author}{Breckon, T.P.}, \bibinfo{year}{2018}.
\newblock \bibinfo{title}{Ganomaly: Semi-supervised anomaly detection via
  adversarial training}.
\newblock \bibinfo{journal}{CoRR} \bibinfo{volume}{abs/1805.06725}.
\newblock \URLprefix \url{http://arxiv.org/abs/1805.06725},
  \href{http://arxiv.org/abs/1805.06725}{{\tt arXiv:1805.06725}}.
%Type = Article
\bibitem[{Ak{\c{c}}ay et~al.(2019)Ak{\c{c}}ay, Abarghouei and Breckon}]{akcay2}
\bibinfo{author}{Ak{\c{c}}ay, S.}, \bibinfo{author}{Abarghouei, A.A.},
  \bibinfo{author}{Breckon, T.P.}, \bibinfo{year}{2019}.
\newblock \bibinfo{title}{Skip-ganomaly: Skip connected and adversarially
  trained encoder-decoder anomaly detection}.
\newblock \bibinfo{journal}{CoRR} \bibinfo{volume}{abs/1901.08954}.
\newblock \URLprefix \url{http://arxiv.org/abs/1901.08954},
  \href{http://arxiv.org/abs/1901.08954}{{\tt arXiv:1901.08954}}.
%Type = Inproceedings
\bibitem[{An and Cho(2015)}]{An2015VariationalAB}
\bibinfo{author}{An, J.}, \bibinfo{author}{Cho, S.}, \bibinfo{year}{2015}.
\newblock \bibinfo{title}{Variational autoencoder based anomaly detection using
  reconstruction probability}.
%Type = Misc
\bibitem[{Arjovsky et~al.(2017)Arjovsky, Chintala and
  Bottou}]{arjovsky2017wasserstein}
\bibinfo{author}{Arjovsky, M.}, \bibinfo{author}{Chintala, S.},
  \bibinfo{author}{Bottou, L.}, \bibinfo{year}{2017}.
\newblock \bibinfo{title}{Wasserstein gan}.
\newblock \href{http://arxiv.org/abs/1701.07875}{{\tt arXiv:1701.07875}}.
%Type = Article
\bibitem[{Bekker and Davis(2018)}]{DBLP:journals/corr/abs-1811-04820}
\bibinfo{author}{Bekker, J.}, \bibinfo{author}{Davis, J.},
  \bibinfo{year}{2018}.
\newblock \bibinfo{title}{Learning from positive and unlabeled data: {A}
  survey}.
\newblock \bibinfo{journal}{CoRR} \bibinfo{volume}{abs/1811.04820}.
\newblock \URLprefix \url{http://arxiv.org/abs/1811.04820},
  \href{http://arxiv.org/abs/1811.04820}{{\tt arXiv:1811.04820}}.
%Type = Article
\bibitem[{Blanchard et~al.(2010)Blanchard, Lee and Scott}]{blanchard10a}
\bibinfo{author}{Blanchard, G.}, \bibinfo{author}{Lee, G.},
  \bibinfo{author}{Scott, C.}, \bibinfo{year}{2010}.
\newblock \bibinfo{title}{Semi-supervised novelty detection}.
\newblock \bibinfo{journal}{Journal of Machine Learning Research}
  \bibinfo{volume}{11}, \bibinfo{pages}{2973--3009}.
\newblock \URLprefix \url{http://jmlr.org/papers/v11/blanchard10a.html}.
%Type = Article
\bibitem[{Chalapathy and Chawla(2019)}]{DBLP:journals/corr/abs-1901-03407}
\bibinfo{author}{Chalapathy, R.}, \bibinfo{author}{Chawla, S.},
  \bibinfo{year}{2019}.
\newblock \bibinfo{title}{Deep learning for anomaly detection: {A} survey}.
\newblock \bibinfo{journal}{CoRR} \bibinfo{volume}{abs/1901.03407}.
\newblock \URLprefix \url{http://arxiv.org/abs/1901.03407},
  \href{http://arxiv.org/abs/1901.03407}{{\tt arXiv:1901.03407}}.
%Type = Article
\bibitem[{Donahue et~al.(2016)Donahue, Kr{\"{a}}henb{\"{u}}hl and
  Darrell}]{DBLP:journals/corr/DonahueKD16}
\bibinfo{author}{Donahue, J.}, \bibinfo{author}{Kr{\"{a}}henb{\"{u}}hl, P.},
  \bibinfo{author}{Darrell, T.}, \bibinfo{year}{2016}.
\newblock \bibinfo{title}{Adversarial feature learning}.
\newblock \bibinfo{journal}{CoRR} \bibinfo{volume}{abs/1605.09782}.
\newblock \URLprefix \url{http://arxiv.org/abs/1605.09782},
  \href{http://arxiv.org/abs/1605.09782}{{\tt arXiv:1605.09782}}.
%Type = Misc
\bibitem[{Dumoulin et~al.(2016)Dumoulin, Belghazi, Poole, Mastropietro, Lamb,
  Arjovsky and Courville}]{dumoulin2016adversarially}
\bibinfo{author}{Dumoulin, V.}, \bibinfo{author}{Belghazi, I.},
  \bibinfo{author}{Poole, B.}, \bibinfo{author}{Mastropietro, O.},
  \bibinfo{author}{Lamb, A.}, \bibinfo{author}{Arjovsky, M.},
  \bibinfo{author}{Courville, A.}, \bibinfo{year}{2016}.
\newblock \bibinfo{title}{Adversarially learned inference}.
\newblock \href{http://arxiv.org/abs/1606.00704}{{\tt arXiv:1606.00704}}.
%Type = Misc
\bibitem[{Goodfellow et~al.(2014)Goodfellow, Pouget-Abadie, Mirza, Xu,
  Warde-Farley, Ozair, Courville and Bengio}]{goodfellow}
\bibinfo{author}{Goodfellow, I.J.}, \bibinfo{author}{Pouget-Abadie, J.},
  \bibinfo{author}{Mirza, M.}, \bibinfo{author}{Xu, B.},
  \bibinfo{author}{Warde-Farley, D.}, \bibinfo{author}{Ozair, S.},
  \bibinfo{author}{Courville, A.}, \bibinfo{author}{Bengio, Y.},
  \bibinfo{year}{2014}.
\newblock \bibinfo{title}{Generative adversarial networks}.
\newblock \href{http://arxiv.org/abs/1406.2661}{{\tt arXiv:1406.2661}}.
%Type = Misc
\bibitem[{Gulrajani et~al.(2017a)Gulrajani, Ahmed, Arjovsky, Dumoulin and
  Courville}]{gulrajani2017improved}
\bibinfo{author}{Gulrajani, I.}, \bibinfo{author}{Ahmed, F.},
  \bibinfo{author}{Arjovsky, M.}, \bibinfo{author}{Dumoulin, V.},
  \bibinfo{author}{Courville, A.}, \bibinfo{year}{2017}a.
\newblock \bibinfo{title}{Improved training of wasserstein gans}.
\newblock \href{http://arxiv.org/abs/1704.00028}{{\tt arXiv:1704.00028}}.
%Type = Article
\bibitem[{Gulrajani et~al.(2017b)Gulrajani, Ahmed, Arjovsky, Dumoulin and
  Courville}]{DBLP:journals/corr/GulrajaniAADC17}
\bibinfo{author}{Gulrajani, I.}, \bibinfo{author}{Ahmed, F.},
  \bibinfo{author}{Arjovsky, M.}, \bibinfo{author}{Dumoulin, V.},
  \bibinfo{author}{Courville, A.C.}, \bibinfo{year}{2017}b.
\newblock \bibinfo{title}{Improved training of wasserstein gans}.
\newblock \bibinfo{journal}{CoRR} \bibinfo{volume}{abs/1704.00028}.
\newblock \URLprefix \url{http://arxiv.org/abs/1704.00028},
  \href{http://arxiv.org/abs/1704.00028}{{\tt arXiv:1704.00028}}.
%Type = Article
\bibitem[{Hou et~al.(2017)Hou, Zhao, Li and
  Chaib{-}draa}]{DBLP:journals/corr/abs-1711-08054}
\bibinfo{author}{Hou, M.}, \bibinfo{author}{Zhao, Q.}, \bibinfo{author}{Li,
  C.}, \bibinfo{author}{Chaib{-}draa, B.}, \bibinfo{year}{2017}.
\newblock \bibinfo{title}{A generative adversarial framework for
  positive-unlabeled classification}.
\newblock \bibinfo{journal}{CoRR} \bibinfo{volume}{abs/1711.08054}.
\newblock \URLprefix \url{http://arxiv.org/abs/1711.08054},
  \href{http://arxiv.org/abs/1711.08054}{{\tt arXiv:1711.08054}}.
%Type = Misc
\bibitem[{Kingma and Welling(2014)}]{kingma2014autoencoding}
\bibinfo{author}{Kingma, D.P.}, \bibinfo{author}{Welling, M.},
  \bibinfo{year}{2014}.
\newblock \bibinfo{title}{Auto-encoding variational bayes}.
\newblock \href{http://arxiv.org/abs/1312.6114}{{\tt arXiv:1312.6114}}.
%Type = Misc
\bibitem[{Kiryo et~al.(2017)Kiryo, Niu, du~Plessis and
  Sugiyama}]{kiryo2017positiveunlabeled}
\bibinfo{author}{Kiryo, R.}, \bibinfo{author}{Niu, G.},
  \bibinfo{author}{du~Plessis, M.C.}, \bibinfo{author}{Sugiyama, M.},
  \bibinfo{year}{2017}.
\newblock \bibinfo{title}{Positive-unlabeled learning with non-negative risk
  estimator}.
\newblock \href{http://arxiv.org/abs/1703.00593}{{\tt arXiv:1703.00593}}.
%Type = Techreport
\bibitem[{Krizhevsky(2009)}]{krizhevsky}
\bibinfo{author}{Krizhevsky, A.}, \bibinfo{year}{2009}.
\newblock \bibinfo{title}{Learning multiple layers of features from tiny
  images}.
\newblock \bibinfo{type}{Technical Report}.
%Type = Article
\bibitem[{LeCun and Cortes(2010)}]{lecun}
\bibinfo{author}{LeCun, Y.}, \bibinfo{author}{Cortes, C.},
  \bibinfo{year}{2010}.
\newblock \bibinfo{title}{{MNIST} handwritten digit database} \URLprefix
  \url{http://yann.lecun.com/exdb/mnist/}.
%Type = Inproceedings
\bibitem[{Lee and Liu(2003)}]{10.5555/3041838.3041895}
\bibinfo{author}{Lee, W.S.}, \bibinfo{author}{Liu, B.}, \bibinfo{year}{2003}.
\newblock \bibinfo{title}{Learning with positive and unlabeled examples using
  weighted logistic regression}, in: \bibinfo{booktitle}{Proceedings of the
  Twentieth International Conference on International Conference on Machine
  Learning}, \bibinfo{publisher}{AAAI Press}. p. \bibinfo{pages}{448–455}.
%Type = Inproceedings
\bibitem[{Li and Liu(2003)}]{10.5555/1630659.1630746}
\bibinfo{author}{Li, X.}, \bibinfo{author}{Liu, B.}, \bibinfo{year}{2003}.
\newblock \bibinfo{title}{Learning to classify texts using positive and
  unlabeled data}, in: \bibinfo{booktitle}{Proceedings of the 18th
  International Joint Conference on Artificial Intelligence},
  \bibinfo{publisher}{Morgan Kaufmann Publishers Inc.}, \bibinfo{address}{San
  Francisco, CA, USA}. p. \bibinfo{pages}{587–592}.
%Type = Inproceedings
\bibitem[{Liu et~al.(2003)Liu, Dai, Li, Lee and Yu}]{10.5555/951949.952139}
\bibinfo{author}{Liu, B.}, \bibinfo{author}{Dai, Y.}, \bibinfo{author}{Li, X.},
  \bibinfo{author}{Lee, W.S.}, \bibinfo{author}{Yu, P.S.},
  \bibinfo{year}{2003}.
\newblock \bibinfo{title}{Building text classifiers using positive and
  unlabeled examples}, in: \bibinfo{booktitle}{Proceedings of the Third IEEE
  International Conference on Data Mining}, \bibinfo{publisher}{IEEE Computer
  Society}, \bibinfo{address}{USA}. p. \bibinfo{pages}{179}.
%Type = Inproceedings
\bibitem[{Liu et~al.(2002)Liu, Lee, Yu and Li}]{Liu02partiallysupervised}
\bibinfo{author}{Liu, B.}, \bibinfo{author}{Lee, W.S.}, \bibinfo{author}{Yu,
  P.S.}, \bibinfo{author}{Li, X.}, \bibinfo{year}{2002}.
\newblock \bibinfo{title}{Partially supervised classification of text
  documents}, pp. \bibinfo{pages}{387--394}.
%Type = Article
\bibitem[{Makhzani et~al.(2015)Makhzani, Shlens, Jaitly and
  Goodfellow}]{DBLP:journals/corr/MakhzaniSJG15}
\bibinfo{author}{Makhzani, A.}, \bibinfo{author}{Shlens, J.},
  \bibinfo{author}{Jaitly, N.}, \bibinfo{author}{Goodfellow, I.J.},
  \bibinfo{year}{2015}.
\newblock \bibinfo{title}{Adversarial autoencoders}.
\newblock \bibinfo{journal}{CoRR} \bibinfo{volume}{abs/1511.05644}.
\newblock \URLprefix \url{http://arxiv.org/abs/1511.05644},
  \href{http://arxiv.org/abs/1511.05644}{{\tt arXiv:1511.05644}}.
%Type = Article
\bibitem[{Metz et~al.(2016)Metz, Poole, Pfau and
  Sohl{-}Dickstein}]{DBLP:journals/corr/MetzPPS16}
\bibinfo{author}{Metz, L.}, \bibinfo{author}{Poole, B.}, \bibinfo{author}{Pfau,
  D.}, \bibinfo{author}{Sohl{-}Dickstein, J.}, \bibinfo{year}{2016}.
\newblock \bibinfo{title}{Unrolled generative adversarial networks}.
\newblock \bibinfo{journal}{CoRR} \bibinfo{volume}{abs/1611.02163}.
\newblock \URLprefix \url{http://arxiv.org/abs/1611.02163},
  \href{http://arxiv.org/abs/1611.02163}{{\tt arXiv:1611.02163}}.
%Type = Article
\bibitem[{Pimentel et~al.(2014)Pimentel, Clifton, Clifton and
  Tarassenko}]{PIMENTEL2014215}
\bibinfo{author}{Pimentel, M.A.}, \bibinfo{author}{Clifton, D.A.},
  \bibinfo{author}{Clifton, L.}, \bibinfo{author}{Tarassenko, L.},
  \bibinfo{year}{2014}.
\newblock \bibinfo{title}{A review of novelty detection}.
\newblock \bibinfo{journal}{Signal Processing} \bibinfo{volume}{99},
  \bibinfo{pages}{215 -- 249}.
\newblock \URLprefix
  \url{http://www.sciencedirect.com/science/article/pii/S016516841300515X},
  \DOIprefix\doi{https://doi.org/10.1016/j.sigpro.2013.12.026}.
%Type = Inproceedings
\bibitem[{Plessis et~al.(2014)Plessis, Niu and
  Sugiyama}]{10.5555/2968826.2968905}
\bibinfo{author}{Plessis, M.C.d.}, \bibinfo{author}{Niu, G.},
  \bibinfo{author}{Sugiyama, M.}, \bibinfo{year}{2014}.
\newblock \bibinfo{title}{Analysis of learning from positive and unlabeled
  data}, in: \bibinfo{booktitle}{Proceedings of the 27th International
  Conference on Neural Information Processing Systems - Volume 1},
  \bibinfo{publisher}{MIT Press}, \bibinfo{address}{Cambridge, MA, USA}. p.
  \bibinfo{pages}{703–711}.
%Type = Inproceedings
\bibitem[{Plessis et~al.(2015)Plessis, Niu and Sugiyama}]{pmlr-v37-plessis15}
\bibinfo{author}{Plessis, M.D.}, \bibinfo{author}{Niu, G.},
  \bibinfo{author}{Sugiyama, M.}, \bibinfo{year}{2015}.
\newblock \bibinfo{title}{Convex formulation for learning from positive and
  unlabeled data}, in: \bibinfo{editor}{Bach, F.}, \bibinfo{editor}{Blei, D.}
  (Eds.), \bibinfo{booktitle}{Proceedings of the 32nd International Conference
  on Machine Learning}, \bibinfo{publisher}{PMLR}, \bibinfo{address}{Lille,
  France}. pp. \bibinfo{pages}{1386--1394}.
\newblock \URLprefix \url{https://proceedings.mlr.press/v37/plessis15.html}.
%Type = Article
\bibitem[{Poole et~al.(2016)Poole, Alemi, Sohl{-}Dickstein and
  Angelova}]{DBLP:journals/corr/PooleASA16}
\bibinfo{author}{Poole, B.}, \bibinfo{author}{Alemi, A.A.},
  \bibinfo{author}{Sohl{-}Dickstein, J.}, \bibinfo{author}{Angelova, A.},
  \bibinfo{year}{2016}.
\newblock \bibinfo{title}{Improved generator objectives for gans}.
\newblock \bibinfo{journal}{CoRR} \bibinfo{volume}{abs/1612.02780}.
\newblock \URLprefix \url{http://arxiv.org/abs/1612.02780},
  \href{http://arxiv.org/abs/1612.02780}{{\tt arXiv:1612.02780}}.
%Type = Misc
\bibitem[{Radford et~al.(2015)Radford, Metz and
  Chintala}]{radford2015unsupervised}
\bibinfo{author}{Radford, A.}, \bibinfo{author}{Metz, L.},
  \bibinfo{author}{Chintala, S.}, \bibinfo{year}{2015}.
\newblock \bibinfo{title}{Unsupervised representation learning with deep
  convolutional generative adversarial networks}.
\newblock \href{http://arxiv.org/abs/1511.06434}{{\tt arXiv:1511.06434}}.
%Type = Article
\bibitem[{Salimans et~al.(2016)Salimans, Goodfellow, Zaremba, Cheung, Radford
  and Chen}]{DBLP:journals/corr/SalimansGZCRC16}
\bibinfo{author}{Salimans, T.}, \bibinfo{author}{Goodfellow, I.J.},
  \bibinfo{author}{Zaremba, W.}, \bibinfo{author}{Cheung, V.},
  \bibinfo{author}{Radford, A.}, \bibinfo{author}{Chen, X.},
  \bibinfo{year}{2016}.
\newblock \bibinfo{title}{Improved techniques for training gans}.
\newblock \bibinfo{journal}{CoRR} \bibinfo{volume}{abs/1606.03498}.
\newblock \URLprefix \url{http://arxiv.org/abs/1606.03498},
  \href{http://arxiv.org/abs/1606.03498}{{\tt arXiv:1606.03498}}.
%Type = Article
\bibitem[{Salimans et~al.(2018)Salimans, Zhang, Radford and
  Metaxas}]{DBLP:journals/corr/abs-1803-05573}
\bibinfo{author}{Salimans, T.}, \bibinfo{author}{Zhang, H.},
  \bibinfo{author}{Radford, A.}, \bibinfo{author}{Metaxas, D.N.},
  \bibinfo{year}{2018}.
\newblock \bibinfo{title}{Improving gans using optimal transport}.
\newblock \bibinfo{journal}{CoRR} \bibinfo{volume}{abs/1803.05573}.
\newblock \URLprefix \url{http://arxiv.org/abs/1803.05573},
  \href{http://arxiv.org/abs/1803.05573}{{\tt arXiv:1803.05573}}.
%Type = Article
\bibitem[{Schlegl et~al.(2017)Schlegl, Seeb{\"{o}}ck, Waldstein,
  Schmidt{-}Erfurth and Langs}]{schlegl}
\bibinfo{author}{Schlegl, T.}, \bibinfo{author}{Seeb{\"{o}}ck, P.},
  \bibinfo{author}{Waldstein, S.M.}, \bibinfo{author}{Schmidt{-}Erfurth, U.},
  \bibinfo{author}{Langs, G.}, \bibinfo{year}{2017}.
\newblock \bibinfo{title}{Unsupervised anomaly detection with generative
  adversarial networks to guide marker discovery}.
\newblock \bibinfo{journal}{CoRR} \bibinfo{volume}{abs/1703.05921}.
\newblock \URLprefix \url{http://arxiv.org/abs/1703.05921},
  \href{http://arxiv.org/abs/1703.05921}{{\tt arXiv:1703.05921}}.
%Type = Article
\bibitem[{Sch\"{o}lkopf et~al.(2001)Sch\"{o}lkopf, Platt, Shawe-Taylor, Smola
  and Williamson}]{scholkopf}
\bibinfo{author}{Sch\"{o}lkopf, B.}, \bibinfo{author}{Platt, J.C.},
  \bibinfo{author}{Shawe-Taylor, J.C.}, \bibinfo{author}{Smola, A.J.},
  \bibinfo{author}{Williamson, R.C.}, \bibinfo{year}{2001}.
\newblock \bibinfo{title}{Estimating the support of a high-dimensional
  distribution}.
\newblock \bibinfo{journal}{Neural Comput.} \bibinfo{volume}{13},
  \bibinfo{pages}{1443–1471}.
\newblock \URLprefix \url{https://doi.org/10.1162/089976601750264965},
  \DOIprefix\doi{10.1162/089976601750264965}.
%Type = Techreport
\bibitem[{Scott and Blanchard(2008)}]{Scott08transductiveanomaly}
\bibinfo{author}{Scott, C.}, \bibinfo{author}{Blanchard, G.},
  \bibinfo{year}{2008}.
\newblock \bibinfo{title}{Transductive anomaly detection}.
\newblock \bibinfo{type}{Technical Report}.
%Type = Article
\bibitem[{Tifrea et~al.(2020)Tifrea, Stavarache and
  Yang}]{DBLP:journals/corr/abs-2012-05825}
\bibinfo{author}{Tifrea, A.}, \bibinfo{author}{Stavarache, E.},
  \bibinfo{author}{Yang, F.}, \bibinfo{year}{2020}.
\newblock \bibinfo{title}{Learn what you can't learn: Regularized ensembles for
  transductive out-of-distribution detection}.
\newblock \bibinfo{journal}{CoRR} \bibinfo{volume}{abs/2012.05825}.
\newblock \URLprefix \url{https://arxiv.org/abs/2012.05825},
  \href{http://arxiv.org/abs/2012.05825}{{\tt arXiv:2012.05825}}.
%Type = Book
\bibitem[{Vapnik(1998)}]{vapnik}
\bibinfo{author}{Vapnik, V.N.}, \bibinfo{year}{1998}.
\newblock \bibinfo{title}{Statistical Learning Theory}.
\newblock \bibinfo{publisher}{Wiley-Interscience}.
%Type = Article
\bibitem[{Yu and Aizawa(2019)}]{DBLP:journals/corr/abs-1908-04951}
\bibinfo{author}{Yu, Q.}, \bibinfo{author}{Aizawa, K.}, \bibinfo{year}{2019}.
\newblock \bibinfo{title}{Unsupervised out-of-distribution detection by maximum
  classifier discrepancy}.
\newblock \bibinfo{journal}{CoRR} \bibinfo{volume}{abs/1908.04951}.
\newblock \URLprefix \url{http://arxiv.org/abs/1908.04951},
  \href{http://arxiv.org/abs/1908.04951}{{\tt arXiv:1908.04951}}.
%Type = Misc
\bibitem[{Zenati et~al.(2018a)Zenati, Foo, Lecouat, Manek and
  Chandrasekhar}]{zenati}
\bibinfo{author}{Zenati, H.}, \bibinfo{author}{Foo, C.S.},
  \bibinfo{author}{Lecouat, B.}, \bibinfo{author}{Manek, G.},
  \bibinfo{author}{Chandrasekhar, V.R.}, \bibinfo{year}{2018}a.
\newblock \bibinfo{title}{Efficient gan-based anomaly detection}.
\newblock \href{http://arxiv.org/abs/1802.06222}{{\tt arXiv:1802.06222}}.
%Type = Article
\bibitem[{Zenati et~al.(2018b)Zenati, Romain, Foo, Lecouat and
  Chandrasekhar}]{zenati2}
\bibinfo{author}{Zenati, H.}, \bibinfo{author}{Romain, M.},
  \bibinfo{author}{Foo, C.S.}, \bibinfo{author}{Lecouat, B.},
  \bibinfo{author}{Chandrasekhar, V.R.}, \bibinfo{year}{2018}b.
\newblock \bibinfo{title}{Adversarially learned anomaly detection}.
\newblock \bibinfo{journal}{CoRR} \bibinfo{volume}{abs/1812.02288}.
\newblock \URLprefix \url{http://arxiv.org/abs/1812.02288},
  \href{http://arxiv.org/abs/1812.02288}{{\tt arXiv:1812.02288}}.

\end{thebibliography}




%%\bibitem[ ()]{}

%%\end{thebibliography}
\end{document}